# Njobvu-AI: An open-source tool for collaborative image labeling and implementation of computer vision models

Jonathan S. Koning[a], Ashwin Subramanian[a], Mazen Alotaibi[a], Cara L. Appel[b,c], Christopher M. Sullivan[a], Thon Chao[d], Lisa Truong[e], Robyn L. Tanguay[e], Pankaj Jaiswal[f], Taal Levi[b], Damon B. Lesmeister[b,c]

*a. Center for Quantitative Life Sciences, Oregon State University, Corvallis, Oregon, USA*
*b. Department of Fisheries, Wildlife, and Conservation Sciences, Oregon State University, Corvallis Oregon, USA*
*c. USDA Forest Service, Pacific Northwest Research Station, Corvallis Oregon, USA*
*d. Department of Integrative Biology, Oregon State University, Corvallis, Oregon, USA*
*e. Department of Environmental and Molecular Toxicology, Oregon State University, Corvallis, Oregon, USA*
*f. Dept. of Botany and Plant Pathology, Oregon State University, Corvallis, Oregon, USA*

## Abstract

Practitioners interested in using computer vision models lack user-friendly and open-source software that combines features to label training data, allow multiple users, train new algorithms, review output, and implement new models. Labeling training data, such as images, is a key step to developing accurate object detection algorithms using computer vision. This step is often not compatible with many cloud-based services for marking or labeling image and video data due to limited internet bandwidth in many regions of the world. Desktop tools are useful for groups working in remote locations, but users often do not have the capability to combine projects developed locally by multiple collaborators. Furthermore, many tools offer features for labeling data or using pre-trained models for classification, but few allow researchers to combine these steps to create and apply custom models. Free, open-source, and user-friendly software that offers a full suite of features (e.g., ability to work locally and online, and train custom models) is desirable to field researchers and conservationists that may have limited coding skills. We developed Njobvu-AI, a free, open-source tool that can be run on both desktop and server hardware using Node.js, allowing users to label data, combine projects for collaboration and review, train custom algorithms, and implement new computer vision models. The name Njobvu-AI (pronounced *N-joh-voo AI*), incorporating the Chichewa word for "elephant," is inspired by a wildlife monitoring program in Malawi that was a primary impetus for the development of this tool and references similarities between the powerful memory of elephants and properties of computer vision models. Code and documentation for this tool are available at https://github.com/sullichrosu/Njobvu-AI/.

## Keywords

computer vision; deep learning; machine learning; object detection; open-source software

## 1. Introduction

Recent developments in computer vision and machine learning techniques have led to an increase in their usage for applied research. Many groups use images, video, and acoustic data for projects across disciplines and are interested in using computer vision and machine learning methods to improve data processing workflows or solve novel problems. Unfortunately, using computer vision and machine learning requires substantial preparation prior to being able to extract desired information from images, video, and audio data.

To start, researchers need to review a subset of their data (e.g., images) and annotate or label them to be used for training. Objects within a set of images are labeled according to classes of interest. This information is then used to train an object-recognition model that can then be used to identify objects in new images. Images and videos are commonly used formats for computer vision models, but audio data may be turned into visual spectrograms and processed using this same pathway.

Currently there are many cloud-based and desktop tools to help users label images and frames of video (e.g., Label Studio [1], OpenLabeling [2], CVAT [3], PowerAI Vision [4]). These tools often have limits around file size, multiple users, and proprietary software, and many have costs for using them. These restrictions limit the usefulness of such tools and offer few options for working outside the cloud or web environment. Additionally, few tools allow users to train a custom computer vision model while retaining control over the training parameters and ownership of their data. Training custom models is therefore limited to users with experience in command-line programming or those using paid services. Here, we present Njobvu-AI (pronounced *N-joh-voo AI*), a free open-source tool that can be used both offline and online.

## 2. Software Description

Njobvu-AI provides a platform that enables users to annotate image datasets in a team environment either via a shared server or as a standalone desktop application, including devices that are not equipped with a reliable internet connection. The ability to annotate, train, and classify images anywhere on any machine while maintaining team cohesion allows for greater flexibility, significantly lowering the barrier to entry into the computer vision side of data science.

### 2.1 Software Architecture

Njobvu-AI is built on Node.js, a JavaScript runtime environment, to run in any web browser and utilizes the templating language EJS to render everything on the server side to maintain performance on the user's machine. All data are uploaded and stored on the server side, which uses SQLite3 to organize the data. There is one global database that stores user information and project metadata. Each project has its own separate database that stores project information such as labels and photos, independent of the user. This compartmentalization allows for fast downloads, easy uploads, and high performance because only smaller databases are accessed during annotation.

Njobvu-AI leverages open-source neural network frameworks to enable in-house computer vision model training and image classification. The two frameworks utilized are Darknet [1] and TensorFlow [2]. Darknet is a computer vision framework predominantly used for object detection, with an emphasis on performance through accuracy and speed using the YOLO algorithm [3]. TensorFlow functions as a more general-purpose framework which supports a large range of applications. Njobvu-AI utilizes Darknet and TensorFlow scripts to train custom models from annotated data, but users can also leverage existing YOLO models to classify new images.

## 2.2 Software Functionalities

### 2.2.1. Creating a user account

When Njobvu-AI is started, the user is brought to the sign-in page. If this is the first time the tool has been used, a profile must be created by using the "Signup" link located at the bottom of the sign-in page. Additional users can be created by using the same sign-up link. The user will be asked to enter details regarding their account setup after selecting the sign-up link.

### 2.2.2. Creating a new project

Projects may be used to organize data by research topic or into smaller batches for labeling. The project setup requires creating an alphanumeric name and description along with uploading a compressed (.zip) file of images. Images should span the full scope of the project with all classes desired to be indexed. Users will also need to provide a comma-separated list of alphanumeric classes that will be used to label objects in the images. New classes can be added inside the project configuration after the project is created if needed.

### 2.2.3. Labeling images

Once a project is created and images are uploaded, users are brought to the home screen where they will be shown all available projects. When the user selects a project, they are provided the list of images that can be labeled using the classes defined at the project's creation. Clicking on the "Label" button to the far right of the list will begin the labeling process. On the Labeling page, a user selects the desired class from the list at the top of the page and then draws a bounding box using click and drag on the image (Figure 1). The border color of the label will correspond to the color of the class. To remove a label, the user simply clicks the box and selects "Delete." There are several shortcut keys to streamline the labeling process—e.g., the "E" key will delete a box. These shortcuts can be found by pressing the "Info" button at the bottom of the page.

Users may find that working with larger images will cause a need to zoom into specific areas to label smaller objects. The tool was developed to maintain proper spatial distances based on zoom level. This means as the user zooms in and out the boxes will retain the size of the box based on the region of the image's bounding box, not the pixels used. The ability to zoom in and out of images properly means that the tool is compatible with even very large images, such as TIFF or LARGE TIFF images from microscopes, CAT (CT) scans, or other equipment. These file formats work seamlessly on the labeling tool in conjunction with the zoom functionality.

Research groups may want to have multiple users review the labeled images, for example, with inexperienced users or images where class identification is uncertain. To accommodate this need, Njobvu-AI has a "Needs Review" button that may be used to flag images that have been labeled and are ready for a secondary person to review the work (Figure 1). Once an image is tagged for review, the project and individual image will be flagged on the home page and project pages (Figure 2). After an image is properly reviewed, the button can be clicked again to remove the tag.

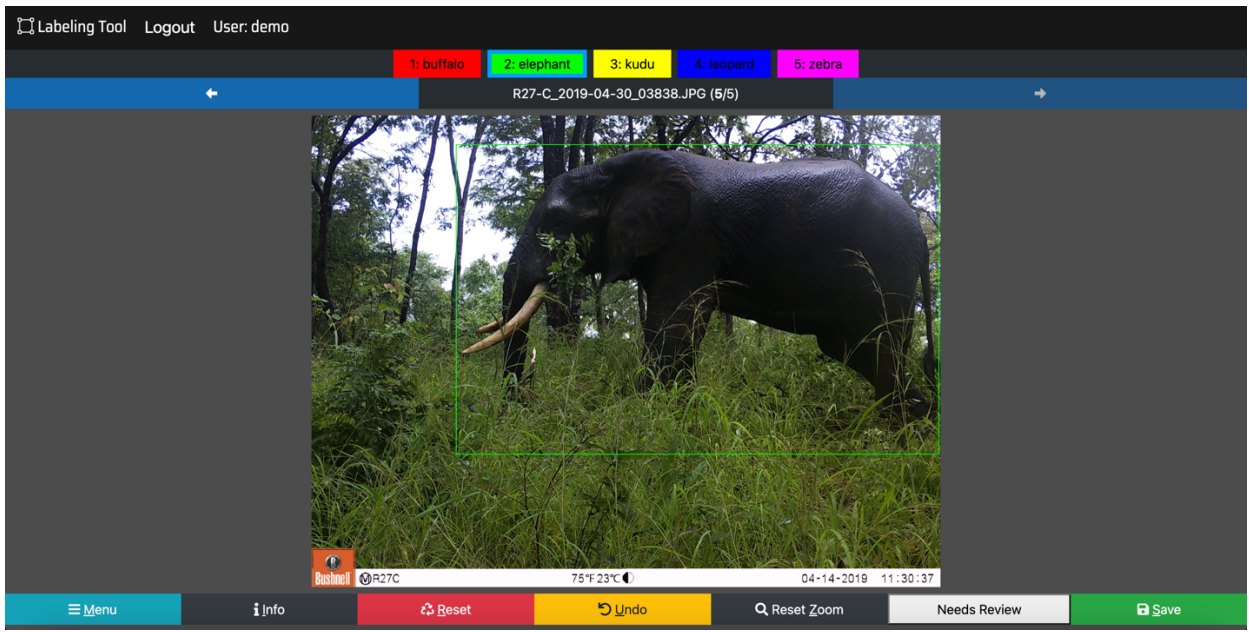

*Figure 1. Labeling page*

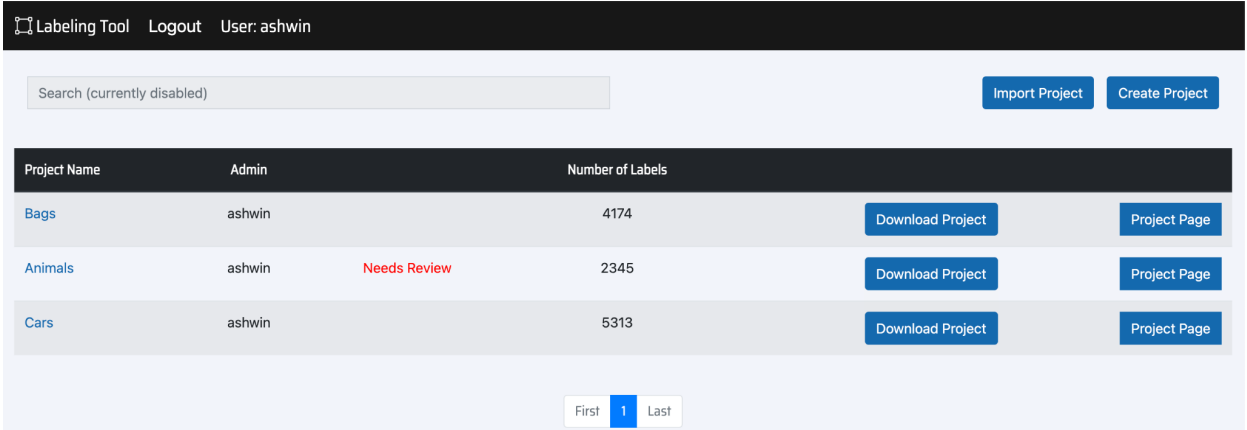

*Figure 2. Project page*

### 2.2.4. Configuring project settings

Each project has a Configuration page with options for custom settings (Figure 3). The configuration page is available to all users; however, the user who created the project is assigned as the administrator, who can enable access for other accounts. Users given access are able to label photos and export results but have limited access to configure the project itself. Only the administrator has full control over the project and has access to permissions such as deleting the project, adding more users to the project, and transferring administrator permissions to another user. Each project has a unique name to distinguish it from other projects along with a project description. These values can be updated, but the name must remain unique from other projects on the same service. The administrator can add new object classes and delete existing classes, after which all previously existing labels of those classes will be removed from the project.

### 2.2.5. Statistics page

Njobvu-AI provides basic statistics about the data for each project on the Statistics page. For each class, the statistics table shows the number of labels across the whole project and the number of images the class appears in. At the top of the page, a percentage is shown, representing the number of images with at least one label out of the total number of images in the project. These statistics give the user a quick overview of the progress of the project as well as a visualization of the balance of labels. This information allows users to estimate which classes their trained model may perform best and which may need more training data.

### 2.2.6. Adding more images

Users may need to increase the amount of data labeled to increase accuracy of the model training. Njobvu-AI has an "Add Images" option on the Configuration page for this purpose. Adding images is similar to creating a new project and requires the user to select a .zip file of images and hit upload. All images are required to have unique filenames, as any images with duplicate filenames within a project will be discarded (Figure 3).

*Figure 3. Configuration page*

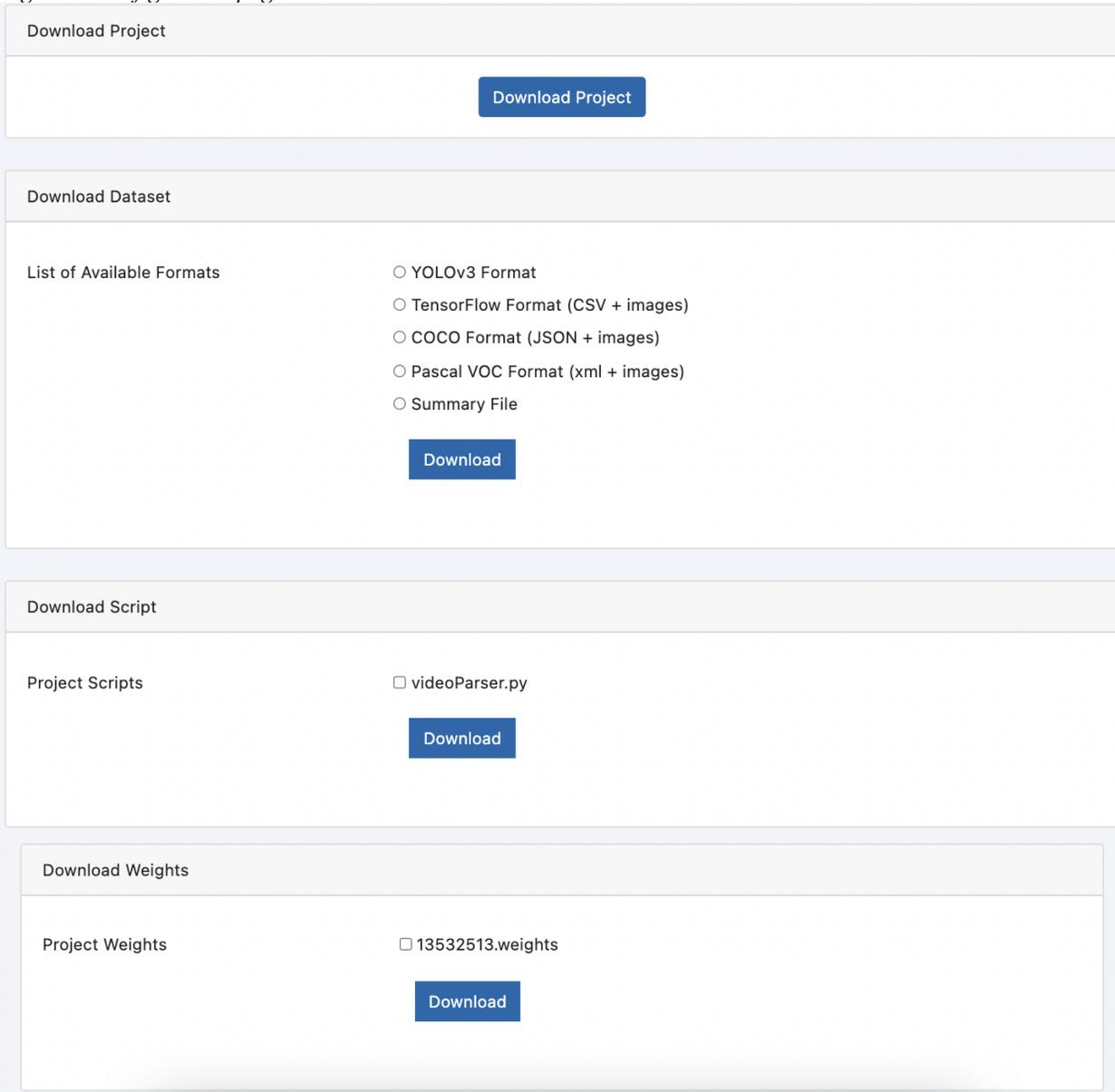

*Figure 4. Download page*

### 2.2.7. Exporting, importing, and merging projects

Users may want to export an entire project to back up their data or move it between machines. On the Download page, users can download the entire project, which will zip all contents (including images) and have it ready for importing into another installation of the tool (Figure 4). Users can also choose to download just the dataset with the labels in either the TensorFlow, YOLO, COCO, or Pascal VOC formats (Figure 4), which can be used to train a computer vision model outside the tool. A summary file can be downloaded to summarize the data in a human-readable format—for example, to count the number of images with each class label, or to associate object labels with other data linked to the image filenames. If any scripts or weight files have been uploaded, they can be downloaded from this page as well. This allows for easy sharing of trained models and base training conditions.

Users can import a downloaded project into a different installation of Njobvu-AI or merge it with a separate project. Existing projects can be imported by clicking the "Import Project" button on the project home page (Figure 2). Merging the data allows groups of users to work on different sets of images and push them back into a single project for review and model training. Merging data can be done by using the Merge Projects option on the main configuration page (Figure 3).

### 2.2.8. Deleting a project

Users can delete a project under the configuration page of the project (Figure 3). Once a project is removed it cannot be brought back unless a backup was created. Deleting a project will also remove associated project files from any stored Darknet paths on the server.

### 2.2.9. Training a user-specified model

Once users have completed labeling their training dataset, they may perform model training using features on either the TensorFlow or YOLO pages. These are two popular machine learning frameworks used in computer vision, allowing users options for training models specifically designed for their data. Users can upload multiple models for simultaneous training and comparison. Having full control of the trained neural network also gives the user confidence their data is not being used to train a larger entity's model.

#### i.  Setting up a training run

*TensorFlow:* On the TensorFlow page (Figure 5), a user can add multiple Python paths that can be used to accommodate and run the user's relevant TensorFlow Python scripts. Most users have multiple versions of Python installed, and the tool allows users to specify the location of the desired Python version. Use of the tool with a remote server negates the need for Python to be installed locally.

*YOLO*: On the YOLO page (Figure 6), users can upload a pre-trained weights file (e.g., YOLOv4) and begin training using their labeled images. Pre-trained models may be downloaded [3] and used for transfer learning, or users may wish to use their own previously trained models for improvement. Users can set a default Darknet file path and have the option of adding and selecting their own list of file paths that point to different installations of Darknet. Users are also given control over their own training specifications (e.g., batch size, subdivisions, width, and height) in order to accommodate systems of various processing power. Users also specify the proportion of the data to be used for training versus withheld for validation.

*Figure 5. TensorFlow page*

*Figure 6. YOLO page*

## ii.    *Running the training*

The length of a training run will differ depending on the amount of labeled data, the training specifications, and the amount of processing power the system has available. Systems that have relatively more powerful Graphics Processing Units (GPUs) can be expected to complete their training in a shorter time period. The status of the training run will be indicated by the Status section on each run (Figure 6). Users can monitor the training progress by clicking the status icon to view the log of the training run. At the bottom of the log file is displayed a rough estimate of the time remaining before the run completes. Once the run is finished, the Status section will change from "Running" to "Done."

### iii.     Generating the weights file

After training, the user can download all files associated with the training using the Download button in the same row. All components including configuration files, log files, and weights files will be bundled together for download. Having the metadata allows for training conditions to be accurately reported and replicated in a new environment. Users will also be able to download any other files created from their scripts.

### 2.2.10. Performing object detection

Once training is completed, users can apply their trained computer vision model to locate and identify objects in new images within Njobvu-AI. This can be accomplished similarly to the process for creating a new project (see 2.2.2). In addition to uploading a .zip file of images, users should select the box titled "Check to Enable Image Classification" and upload a .zip file containing the model weights file and setup files used to generate the weights file from the trained model (2.2.9). Njobvu-AI will then upload, process, and label the set of new images. This will create a new project in which each image contains bounding box(es) around every predicted object within it. The resulting labels around each object will behave exactly as if the user had labeled each image manually, and users can review and modify the predicted labels if needed.

### 2.2.11. Performing validation on classified projects

Users can navigate to Validation Mode by selecting the Training Mode button on the main page and vice versa. Validation mode can be used to assess the performance of trained model on an unlabeled dataset as well as enable faster correcting of labeled photos. This mode offers specific features for correcting the labels in a dataset as well as viewing the results, such as sorting by class, bulk class changes, and individual photo statistics.

### 2.2.12. Online and offline modes

Njobvu-AI is built on Node.js, so users can easily bring the system online using any web server hosting Node.js services. Njobvu-AI can be put into a cloud service and accessed from around the world if needed. Since we created Njobvu-AI as a small, contained package system, users can also install Node.js locally and run the tool on a desktop machine. We have successfully run the system on Windows, Mac, and Linux operating systems as well as x86 and POWER architectures without issues.

## 3. Impact

All current labeling tools attempt to achieve the same basic principles, to generate a dataset of manually annotated images to train a neural network. A considerable limitation of many tools is the inability to exist anywhere other than a local environment. This hampers the ability of researchers to access their data from anywhere in the world and limits the scope of the project to a single machine, restricting collaboration.

Some available tools for annotation and model training, such as IBM's PowerAI Vision [6], overcome this limitation by allowing for local and web environments with users and administrators for each project. PowerAI Vision also has a user-friendly interface that lowers the learning curve. However, these enhanced features come at the cost of a substantial annual paywall. In addition to the significant annual costs, when using PowerAI Vision's proprietary algorithms to train a dataset, the resulting model is restricted to the PowerAI software and is inaccessible once subscription lapses. Many researchers will be unable to afford labeling tools such as PowerAI and in turn the software becomes limited to those who have the budget to afford it. Because Njobvu-AI is open source, there is little-to-no barrier to entry for researchers and conservationists to have access to a high-quality labeling and model training tool. Even outside of research, people with any interest in machine learning can use Njobvu-AI.

## 4. Conclusions

Njobvu-AI is a new browser-based tool for image annotation, model training, and object detection that aims to fill current gaps in available tools. Our tool is designed to work as both a cross-platform standalone app and a cloud-based service that empowers teams to work together. With the ability to download and merge projects, group members who are working at the edge can easily upload their progress to the cloud or transfer their project data to other local machines, keeping the rest of the team up to date with the latest images and labels. To help maintain high-quality data, Njobvu-AI supports a review button which flags the specified image for team members to review, which allows for error checking and expert review if needed. After labeling is complete, users can train their model from within the tool or download their dataset in either TensorFlow or YOLO Darknet format. When training from within the tool, users can upload a pre-trained model and select the Python installation to run it with. Finally, users can apply trained models to classify new data within the tool and use the Validation mode to review object detection predictions. Few other available tools provide a complete pipeline from tagging to inference using custom-trained models. This gives the user complete control and customizability over their neural network algorithms. Njobvu-AI is completely open source, allowing users to add custom formats and guarantee privacy.

## Acknowledgements

Data used for developing this tool were acquired through cooperation with African Parks and U.S. Forest Service International Programs. Funding was provided by USDA Forest Service, Pacific Northwest Research Station. The findings and conclusions in this publication are those of the authors and should not be construed to represent any official U.S. Department of Agriculture or U.S. Government determination or policy. The use of trade or firm names in this publication is for reader information and does not imply endorsement by the U.S. Government of any product or service. The research was also partially supported by the National Institute of Environmental Health Sciences of the National Institutes of Health under Award P30 ES030287. The content is solely the responsibility of the authors and does not necessarily represent the official views of the National Institutes of Health.